\documentclass[letterpaper, 10 pt, conference]{ieeeconf}
\IEEEoverridecommandlockouts                              

\usepackage{graphicx}
\usepackage{amsmath,amssymb,amsfonts}
\usepackage{booktabs}
\usepackage{algorithm}
\usepackage{algpseudocode}
\usepackage{mathtools}
\usepackage{cite}
\usepackage{subcaption}
\usepackage{hyperref}
\usepackage{xcolor}

\newcommand{\bphi}{\boldsymbol{\phi}}
\newcommand{\beps}{\boldsymbol{\epsilon}}

\newcommand{\bX}{\textbf{\emph{X}}}

\newcommand{\bt}{\textbf{\emph{t}}}

\newcommand{\bolr}{\textbf{\emph{r}}}

\newcommand{\bu}{\textbf{\emph{u}}}

\newcommand{\bp}{\textbf{\emph{p}}}

\newcommand{\bA}{\textbf{\emph{A}}}
\newcommand{\bC}{\textbf{\emph{C}}}
\newcommand{\bD}{\textbf{\emph{D}}}

\newcommand{\bI}{\textbf{\emph{I}}}
\newcommand{\bP}{\textbf{\emph{P}}}
\newcommand{\bK}{\textbf{\emph{K}}}

\newcommand{\bR}{\textbf{\emph{R}}}

\newcommand{\bk}{\textbf{\emph{k}}}
\newcommand{\bB}{\textbf{\emph{B}}}
\newcommand{\bT}{\textbf{\emph{T}}}
\newcommand{\bW}{\textbf{\emph{W}}}

\newcommand{\bM}{\textbf{\emph{M}}}

\newcommand{\bS}{\textbf{\emph{S}}}
\newcommand{\bU}{\textbf{\emph{U}}}
\newcommand{\bV}{\textbf{\emph{V}}}
\newcommand{\eg}{e.g.,~}

\newcommand{\bSigma}{\boldsymbol{\Sigma}}

\title{\LARGE \bf
Optimal DLT-based Solutions for the Perspective-n-Point
}

\author{S\'{e}bastien Henry$^{1}$ and John A. Christian$^{1}$
\thanks{$^{1}$S\'{e}bastien Henry and John A. Christian are with Guggenheim School of Aerospace Engineering,  Georgia Institute of Technology,  30332,  USA.
        {\tt\small seb.henry@gatech.edu}, {\tt\small john.a.christian@gatech.edu}}%
}

\begin{document}

\maketitle

\begin{abstract}
We propose a modified normalized direct linear transform (DLT) algorithm for solving the perspective-n-point (PnP) problem with much better behavior than the conventional DLT. The modification consists of analytically weighting the different measurements in the linear system with a negligible increase in computational load. Our approach exhibits clear improvements---in both performance and runtime---when compared to popular methods such as EPnP, CPnP, RPnP, and OPnP. Our new non-iterative solution approaches that of the true optimal found via Gauss-Newton optimization, but at a fraction of the computational cost. Our optimal DLT (\texttt{oDLT}) implementation, as well as the experiments, are released in open source. \footnote{Open source code will be made available at \href{https://github.com/WaterEveryDay/oDLT}{https://github.com/WaterEveryDay/oDLT}. This work has been submitted to the IEEE for possible publication. Copyright may be transferred without notice, after which this version may no longer be accessible. }
\end{abstract}


\section{INTRODUCTION}
Position and attitude (i.e., pose) estimation from points in correspondence is a recurring and pervasive problem in computer vision. Especially interesting is the task of finding the pose using only a single image from a monocular camera. Assuming that the image contains the perspective projection of $n$ world points, we refer to this as the perspective-$n$-point (PnP) problem. At least three points are required for a solution to exist, leading to the minimal P3P problem for which solutions have existed for over a century \cite{Grunert:1841_P3P, Haralick:1991_P3P,Gao:2003_P3P}. Although tailored solutions exist for other small problems (e.g., P4P \cite{Quan:1999_PnP}), modern computer vision problems generally have a large number of points (i.e., $n\gg3$) and the general PnP problem is of principal interest. A recent summary of these methods can be found in Ref.~\cite{li:2023_summary}. 

Most statistically optimal PnP solutions in the extant literature are iterative. Iterative solutions are usually slower than non-iterative solutions since they (1) often require the use of an initial estimate of the camera pose and (2) solve a problem repeatedly (that is, iteratively) until some condition is met. We accept this computational burden since the iterative methods can produce the optimal solution of the full problem. Standard algorithms for solving non-linear least squares problems, such as Levenberg-Marquardt (LM) or Gauss-Newton (GN), iteratively refine an initial estimate---thus requiring \textit{a priori} information to generate that initial guess \cite{hartley_zisserman_2004}. 
Specialized PnP solutions are conceptually similar in structure \cite{Lu:2000_LHM}. 
Iterative solutions are particularly appealing for problems that are not necessarily well behaved in terms of noise or geometry and are of small size.

\begin{figure}[t!]
    \centering
        \hfill
        \begin{subfigure}{0.48\linewidth}
        \centering
        \includegraphics[width=1.\linewidth]{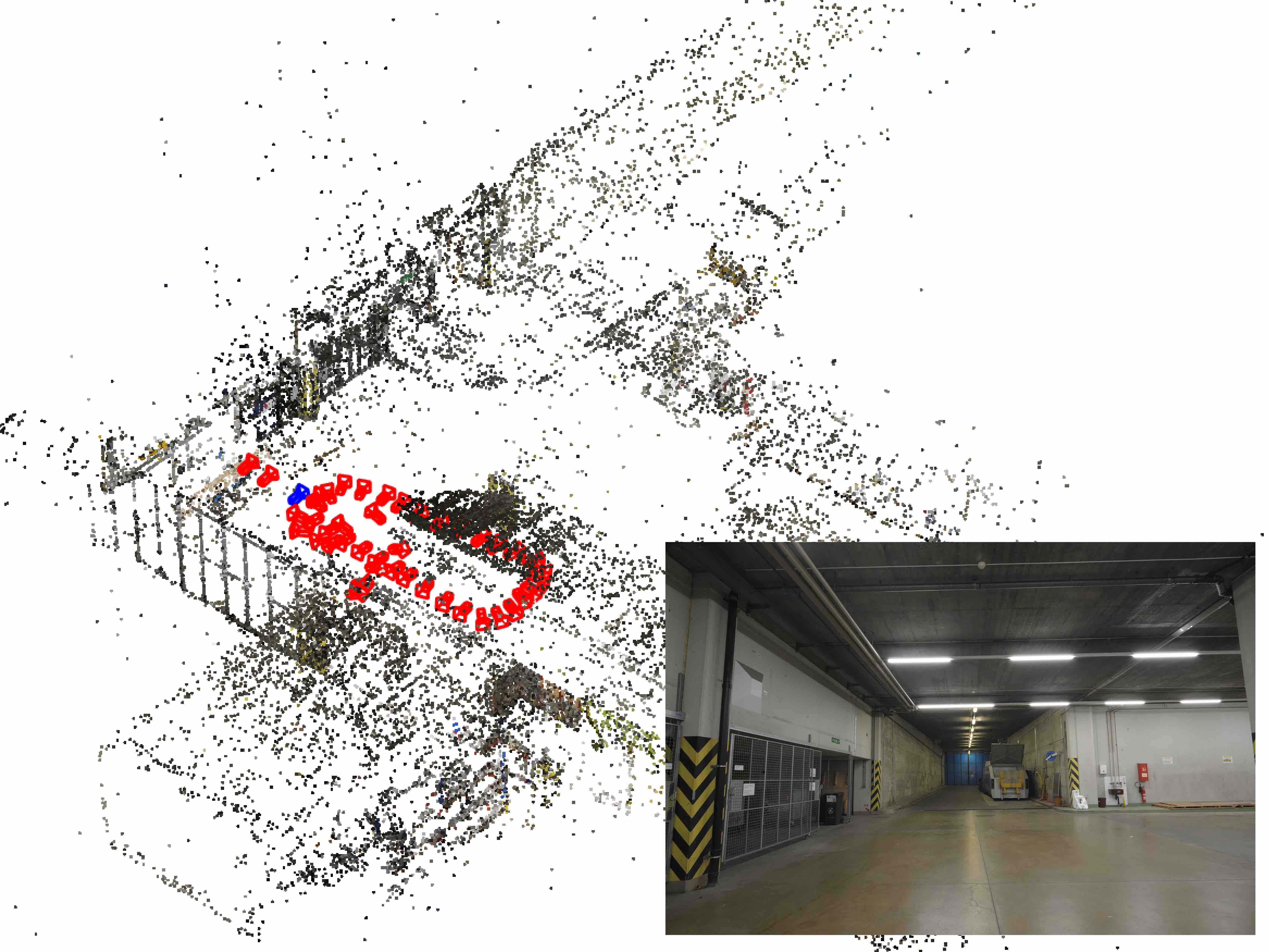}
        \caption{Delivery Area.}
        \end{subfigure}
        \hfill
        \begin{subfigure}{0.48\linewidth}
        \includegraphics[width=1.\linewidth]{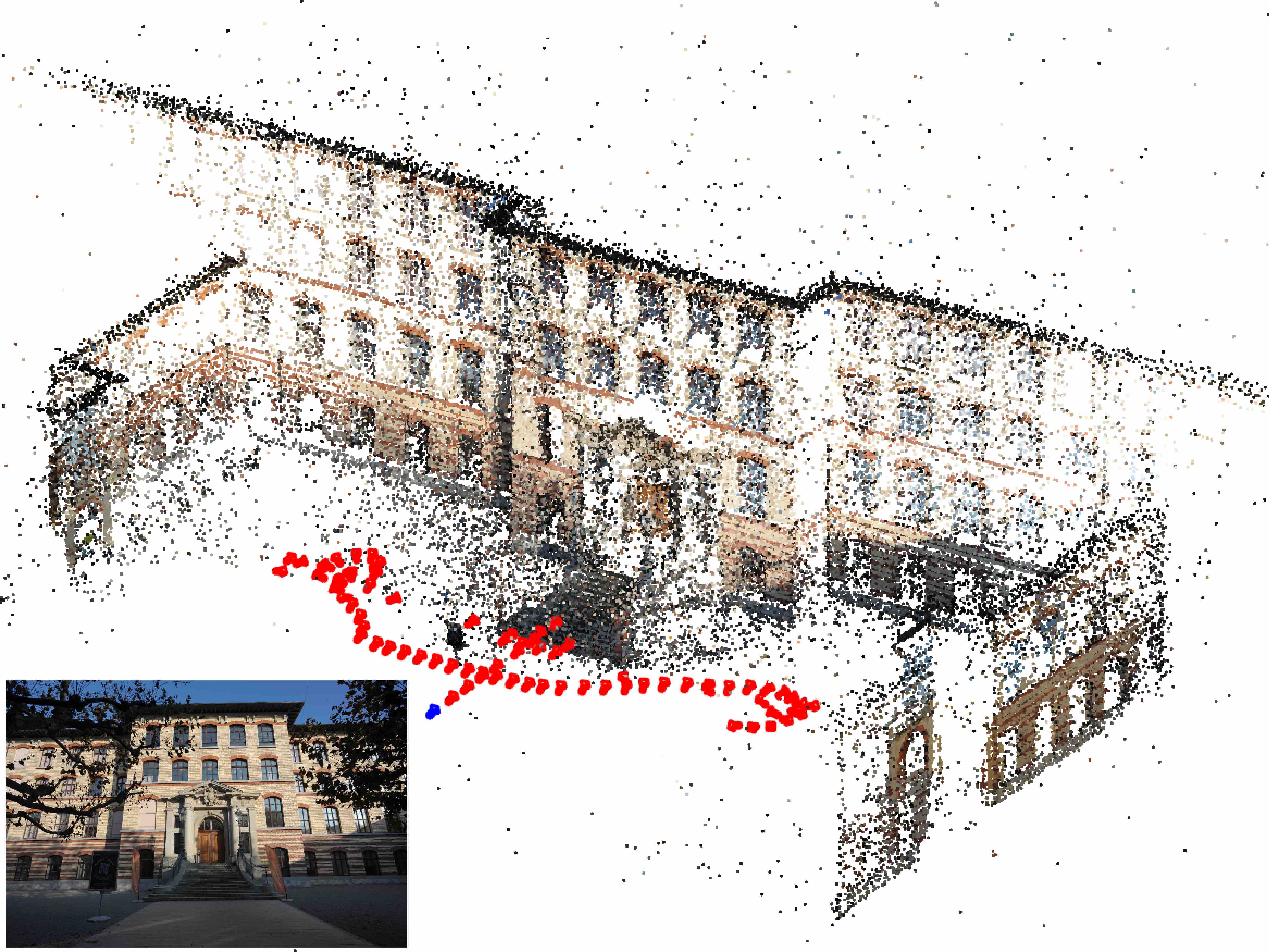}
        \caption{Facade.}
        \end{subfigure}
        \hfill

        \hfill
        \begin{subfigure}{0.48\linewidth}
        \centering
        \includegraphics[width=1.\linewidth]{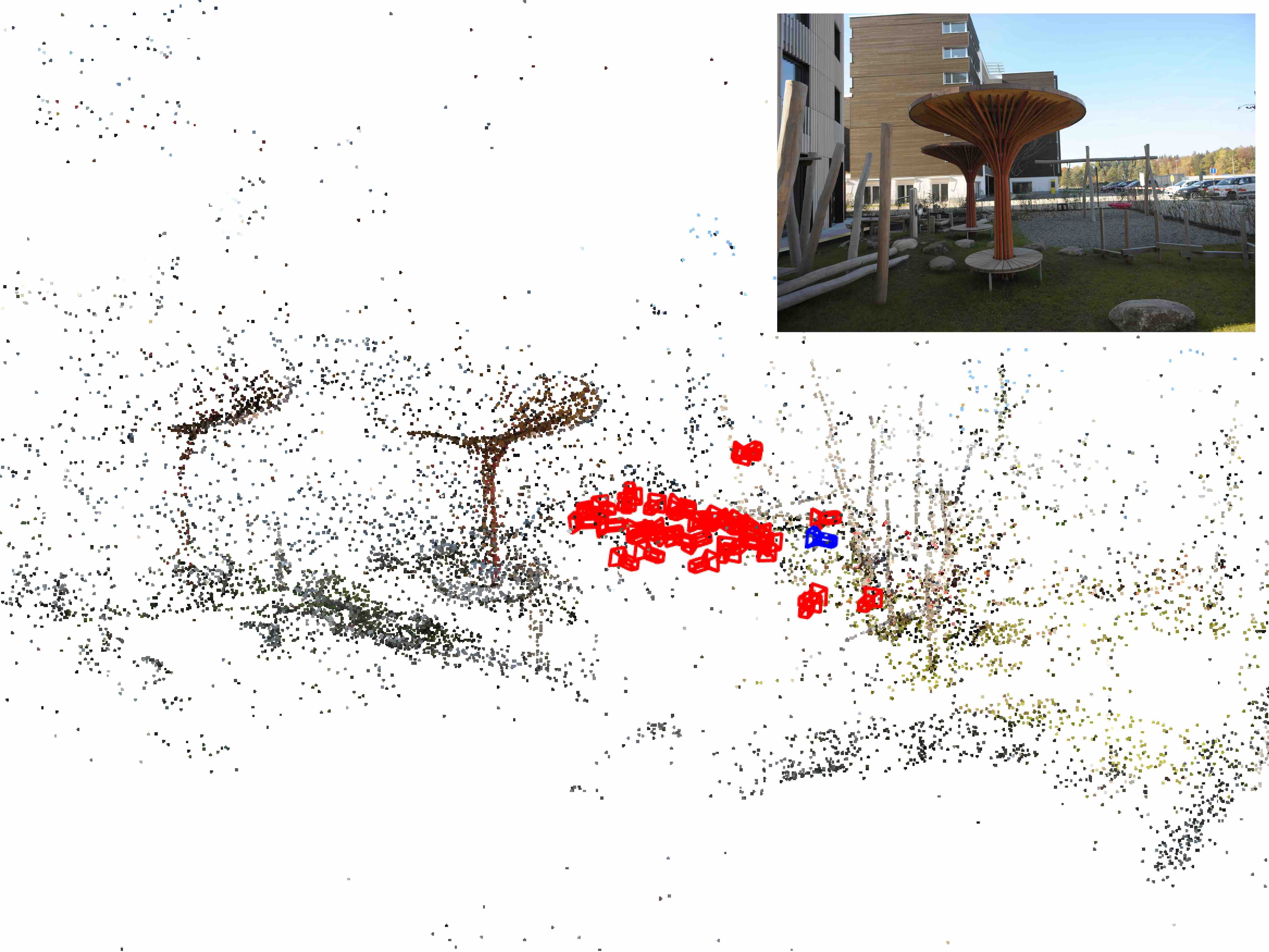}
        \caption{Playground.}
        \end{subfigure}
        \hfill
        \begin{subfigure}{0.48\linewidth}
        \includegraphics[width=1.\linewidth]{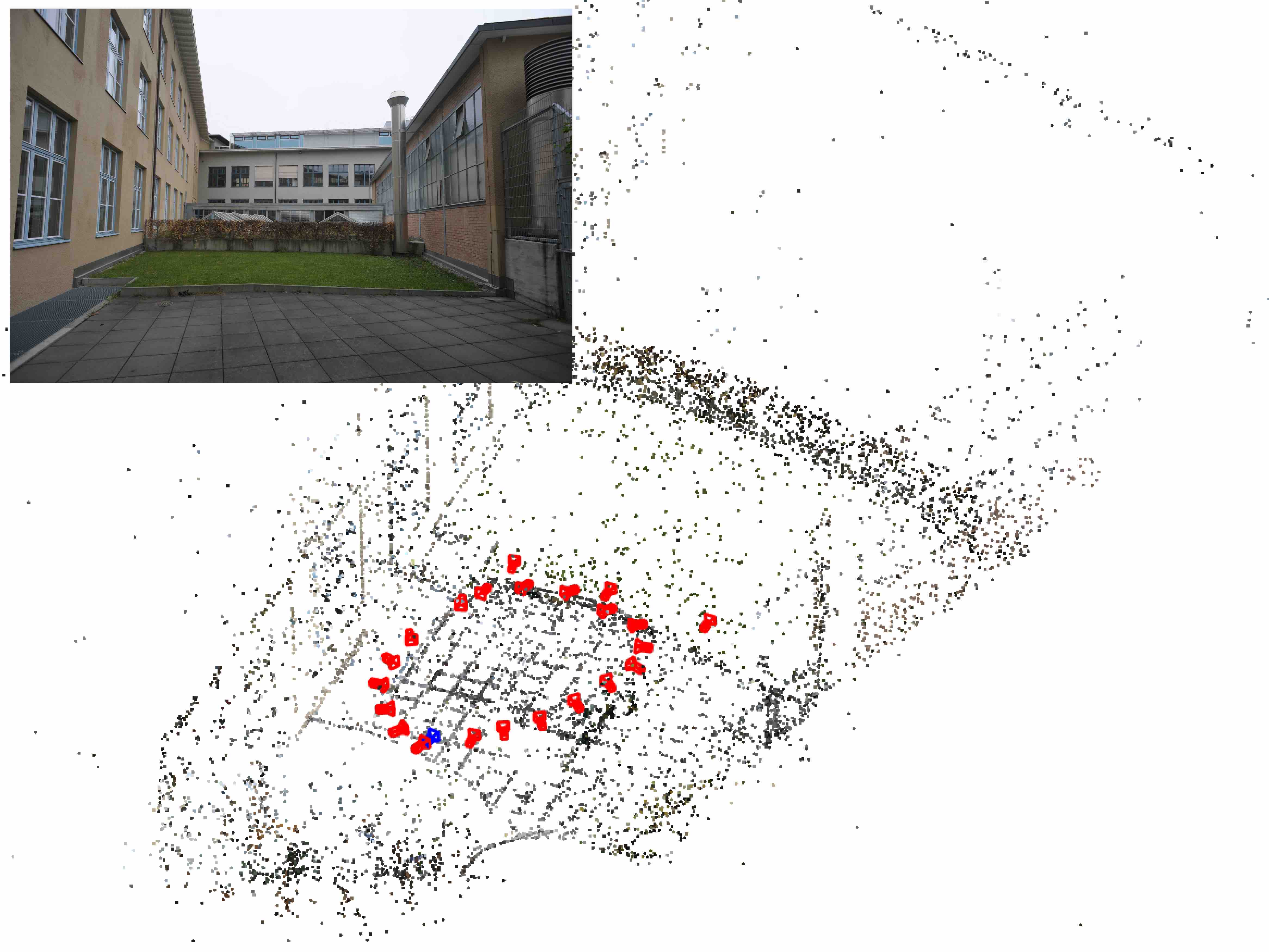}
        \caption{Terrace.}
        \end{subfigure}
        \hfill
    \caption{\texttt{oDLT+LOST} camera pose estimations on some ETH3D training sets. As a visual help, the camera is drawn in blue when it corresponds to the shown image.}
    \label{fig:eth_visuals}
\end{figure}

Non-iterative solutions to the PnP problem are faster than iterative solutions, but this speed usually comes at the expense of statistical optimality (i.e., pose estimation performance). Early works managed to find solutions with complexity ranging from $\mathcal{O}(n^8)$ \cite{Ansar:2003} down to $\mathcal{O}(n^5)$ \cite{Quan:1999_PnP} and $\mathcal{O}(n^2)$ \cite{Fiore:2001_PnP}. More recently, a number of non-iterative algorithms find a solution in $\mathcal{O}(n)$ \cite{lepetit:2009_epnp, Zheng:2013_opnp}. Specialized methods exist for a number of cases, such as algorithms that work best in the planar case / singular case \cite{li:2012_RPnP}, that provide good solutions from $n\geq 3$ \cite{Hesch:2011_DLS, Zheng:2013_opnp}, that remove bias from the solution \cite{Zeng:2023_CPnP}, that focus on outlier rejection \cite{Ferraz:2014_REPPnP}, that ensure a positive depth in the solution \cite{li:2023_summary}, or that focus on global optimality \cite{Zheng:2013_opnp, terzakis:2020_SQPnP}. Most methods try to minimize the geometric error that provides the maximum likelihood estimator when the 3D points are perfectly known. However, when the structure is imperfect, one can also take the uncertainties into account \cite{ferraz:2014_uncertainty_pnp, Urban:2016_MLPnP, Vakhitov:2021_EPNPU}.

The direct linear transform (DLT)  \cite{AbdelAziz_1971:DLT, Sutherland:1974_DLT, hartley_zisserman_2004} was one of the first $\mathcal{O}(n)$ solutions to the PnP problem. However, the classical DLT is known to have limited performance due to the fact that it estimates a matrix that implicitly contains the calibration matrix. Even when clamping the known calibration matrix to the solution, the classical DLT is often worse when compared to other PnP methods. The behavior of the DLT, however, can be significantly improved by normalizing the data beforehand \cite{hartley_zisserman_2004}. Another inherent problem to the DLT is that the solution is not strictly constrained to $SE(3)$, and thus an extra step is necessary to orthonormalize the rotation matrix and recover the scale. Despite these drawbacks, the DLT remains a computationally light solution that is easy to understand and to implement.

This paper contributes to the state-of-the-art by improving the performance of the DLT for PnP, by leveraging ideas from optimal triangulation theory \cite{Henry:2022_LOST} and star-based attitude determination \cite{Christian:2021_starid}. We show that the DLT can be weighted in a maximum likelihood sense if there is \textit{a priori} information on the camera matrix. The \textit{a priori} information can easily be obtained by a smaller-sized normalized DLT, RANSAC \cite{Fischler:1981_RANSAC} --- with the benefit of providing inliers---, or other inexpensive methods. Furthermore, we leverage the covariance information of the solution to weigh the orthogonal Procrustes problem \cite{schonemann:1966_generalized} and retrieve an optimal $SO(3)$ rotation. While this two-shot approach comes at little extra computational cost for large $n$, it provides a dramatic improvement to the normalized DLT. Indeed, we find that our optimized DLT (\texttt{oDLT}) provides pose performance similar to the iterative methods but at the computational expense of the DLT.

Section \ref{sec:normalized_DLT} reviews the basic theory of the normalized DLT for camera pose estimation. Then, Section \ref{sec:oDLT} presents an approach to transform the DLT system into a maximum likelihood estimator. Finally, in Section \ref{sec:experiments}, we test the DLT on simulated data as well as real benchmarks \cite{schoeps:2017_ETH3D, Driver:2023_AstroVision}.

\section{STANDARD NORMALIZED DLT} \label{sec:normalized_DLT}
Consider a known 3D point expressed in the world coordinate frame, $\bp_i \in \mathbb{R}^{3\times1}$ and its homogeneous coordinate $\bar{\bp}_i = \begin{bmatrix}
    \bp_i^T & 1
\end{bmatrix}^T$. 
The corresponding 2D pixel measurement is $\bu_i =  \begin{bmatrix}
    u_i & v_i
\end{bmatrix}^T$, or $\bar{\bu}_i = \begin{bmatrix}
        \bu_i^T & 1
    \end{bmatrix}^T$ expressed in homogeneous coordinates. The pixel measurements are related to the 3D points using the camera projection matrix $\bP \in \mathbb{R}^{3\times4}$
\begin{equation} \label{eq:projection}
    \bar{\bu}_i = \frac{\bK \bR (\bp_i - \bolr)}{\bk^T \bK \bR (\bp_i - \bolr)} = \frac{\bP \bar{\bp}_i}{\bk^T \bP \bar{\bp}_i},
\end{equation}
where $\bk^T = [0, 0, 1]$.
The camera projection matrix can be decomposed as
\begin{equation} \label{eq:projmat}
    \bP = \bK \bR \left[ \bI_{3\times3}, -\bolr \right] = \bK \left[ \bR , \bt \right]
\end{equation}
where $\bK$ is the camera calibration matrix (intrinsic parameters), $\bR$ is the rotation matrix from the world coordinate system to the camera coordinate system and $\bolr$ is the position of the camera center in the world coordinate system. The purpose of the PnP problem is to estimate the extrinsic parameters $\bR$ and $\bolr$ for a calibrated camera ($\bK$ is known) using measurements $\bu_i$ corresponding to real 3D points $\bp_i$. The DLT system may be built by noting the collinearity between $\bar{\bu}_i$ and $\bP \bar{\bp}_i$, such that their cross product is zero,
\begin{equation}
    \label{eq:DLT_cross_product}
    [\bar{\bu}_i \times] \bP \bar{\bp}_i = \mathbf{0}_{3\times1}.
\end{equation}
We observe this equation to be linear in the unknown $\bP$. Consequently, making use of the Kronecker product $\otimes$, and using the fact that $\text{vec}\left(\bA \bX \bB\right) = \left(\bB^T \otimes \bA \right) \text{vec}\left(\bX\right)$,  we may factor out the vectorized $\bP$ and rewrite Eq.~\eqref{eq:DLT_cross_product} as
\begin{equation}
    \label{eq:geometric_constraint}
    \left( \bar{\bp}^T_i \otimes [\bar{\bu}_i \times] \right) \text{vec}(\bP) = \bA_i \text{vec}(\bP) = \mathbf{0}_{3\times1}.
\end{equation}
where $\bA_i = \bar{\bp}^T_i \otimes [\bar{\bu}_i \times] $ is a $3 \times 12$ matrix consisting only of values known \emph{a priori}.

The vec operator will be used for covariance propagation in the next section. Since $[\bar{\bu}_i \times]$ is only rank two, the third row of $\bA_i$ is a linear combination of the first two rows and does not contain any unique information, thus it can be eliminated to speed up computations. Hence we proceed with the matrix $\bS \bA_i$, where $\bS = [\bI_{2\times2}, \mathbf{0}_{2\times1}]$ removes the third redundant row. The camera matrix comprises 12 elements, minus 1 degree of freedom for scale. The 11 degrees of freedom can be seen as 6 for the pose plus 5 for the calibration. Since each point adds two degrees of freedom, 6 points are in general necessary to solve the PnP with the DLT. For $n\geq 6$ measurements, we stack Eq.~\ref{eq:geometric_constraint} for each of the $n$ measurements to form the system
\begin{equation} \label{eq:DLT_to_solve}
    \bA \text{vec}(\bP) = \begin{bmatrix}
        \bS \bA_1 \\ \bS \bA_2 \\ \vdots \\ \bS \bA_n,
    \end{bmatrix} \text{vec}(\bP) = \mathbf{0}_{2n\times1}
\end{equation}
where $\bA \in \mathbb{R}^{2 n \times 12}$.

The least-squares solution to this homogeneous system is the null space of A, which corresponds to the one-dimensional subspace of the smallest singular value of $\bA$. Thus it may be found in the last column vector of $\bV$ from the singular value decomposition (SVD)
\begin{equation} \label{eq:svdsoln}
    \bA = \bU \bD \bV^T. 
\end{equation}
The solution to Eq.~\ref{eq:DLT_to_solve} gives the camera matrix $\bP$ up to an arbitrary scale. The scale can be recovered by recognizing that $\det(\bR) = 1$. Importantly, the scale is not the only unsatisfied constraint, there is also no guarantee that the solution for $\bR$ from Eq.~\ref{eq:svdsoln} is an orthonormal matrix. The closest orthonormal matrix can be found by solving the orthogonal Procrustes problem \cite{schonemann:1966_generalized}, which involves another SVD. 

The result from the DLT is non-invariant under a similarity transformation (as noted by Hartley and Zisserman~\cite{hartley_zisserman_2004}), and solving the PnP with the DLT behaves much better if the data is recentered and scaled beforehand. In particular, the pixel points $\bu_i$ should be recentered such that their mean is at $[0, 0]^T$ and then scaled such that their average distance from the origin is $\sqrt{2}$. As a consequence, a similarity transformation $\bT_u\in \mathbb{R}^{3\times3}$ can be computed, such that the normalized coordinates are
\begin{equation} \label{eq:normalize_meas}
    \tilde{\bar{\bu}}_i = \bT_u \bar{\bu}_i.
\end{equation}
Similarly, the 3D points need to also be recentered with a mean of $[0, 0, 0]^T$ and scaled for an average distance from the origin $\sqrt{3}$. For that we find the similarity transformation $\bT_p\in \mathbb{R}^{4\times4}$ such that the normalized points are
\begin{equation}  \label{eq:normalize_points}
    \tilde{\bar{\bp}}_i = \bT_p \bar{\bp}_i.
\end{equation}
Thus the system in \ref{eq:geometric_constraint} and \ref{eq:DLT_to_solve} should be built using $\tilde{\bar{\bu}}_i$ and $\tilde{\bar{\bp}}_i$ to yield the solution $\tilde{\bP}$. The solution in the original, un-normalized space can be found via
\begin{equation} \label{eq:un-normalize}
    \bP = \bT_u^{-1} \tilde{\bP} \bT_p .
\end{equation}

\section{OPTIMAL NORMALIZED DLT} \label{sec:oDLT}
Noisy measurements cause the constraint in Eq.~\ref{eq:geometric_constraint} to not be fully satisfied. Instead, the matrix-vector product gives a residual that is the \emph{algebraic} error
\begin{equation} 
\label{eq:eps_i}
    \beps_i = \bA_i \text{vec}(\bP) \neq \textbf{0}_{3 \times 1}.
\end{equation}
It is the algebraic error that is minimized by the standard DLT, which makes no distinction about the quality of each separate measurement. However, prior works have demonstrated that minimizing the \emph{reprojection} error leads to the maximum likelihood solution~\cite{hartley_zisserman_2004}. We thus seek to find $\bP$ that minimizes the displacement of the measurements on the image,
\begin{equation}
    \text{min}_\bP \sum_{i=1}^n d(\bar{\bu}_i, \bP \bar{\bp_i}).
\end{equation}
Conventionally, the DLT is used as a first approximate solution and then an iterative correction step (e.g., using the Levenberg-Marquadt algorithm) is used to converge to the optimal solution. 

In order to find an alternate maximum likelihood estimator (MLE), let us consider the covariance of the residuals, $\bSigma_{\beps_i}$. We formulate the standard MLE for Gaussian noise
\begin{equation}
    \min_\bP \sum_i \beps_i^T \bSigma_{\beps_i}^\dagger \beps_i = \sum_i \text{vec}(\bP)^T \bA_i^T \bSigma_{\beps_i}^\dagger \bA_i \text{vec}(\bP),
\end{equation}
where we consider the pseudoinverse ($\dagger$) because $\bSigma_{\beps_i}$ is not full rank, but its null space aligns with that of $\text{null}(\bA_i^T)$. From the first differential condition, we require that
\begin{equation}
    2 \sum_i \text{vec}(\bP)^T \bA_i^T \bSigma_{\beps_i}^\dagger \bA_i = \mathbf{0}_{1\times12},
\end{equation}
or, equivalently, 
\begin{equation}
    \left(\sum_i \bA_i^T \bSigma_{\beps_i}^\dagger \bA_i\right) \text{vec}(\bP) = \mathbf{0}_{12\times1}.
\end{equation}

Replacing $\bSigma_{\beps_i}$ by $\bI_{3\times3}$ yields to the same solution as the DLT. The linear relation between the residual and the pixel measurement can be found as
\begin{equation}
\begin{aligned}
    \beps_i &= \bA_i \text{vec}(\bP) = \left[ \bar{\bu}_i \times \right] \bP \bar{\bp}_i = - \left[  \bP \bar{\bp}_i\times \right]  \bar{\bu}_i
\end{aligned}
\end{equation}
Thus, the covariance of $\beps_i$ is related to that of $\bar{\bu}_i$ and $\bar{\bp}_i$ by the following expression:
\begin{equation}
    \bSigma_{\beps_i} = - \left[ \bP \bar{\bp}_i \times \right] \bSigma_{\bar{\bu}_i} \left[ \bP \bar{\bp}_i \times \right] - \left[ \bar{\bu}_i \times \right]\bP \bSigma_{\bar{\bp}_i} \bP^T\left[ \bar{\bu}_i \times \right],
\end{equation}
which by using $\left[ \bP \bar{\bp}_i \times \right] = \bk^T \bP \bar{\bp}_i \left[ \bar{\bu}_i \times \right]$ (Eq. \ref{eq:projection}), yields
\begin{equation} \label{eq:beps_2D+3D}
    \bSigma_{\beps_i} =  - \left[ \bar{\bu}_i \times \right] \left(\left(\bk^T \bP \bar{\bp}_i \right)^2 \bSigma_{\bar{\bu}_i} +  \bP \bSigma_{\bar{\bp}_i} \bP^T \right)\left[ \bar{\bu}_i \times \right].
\end{equation}
This expression in Eq.~\ref{eq:beps_2D+3D} is very similar to the one obtained for triangulation in Ref.~\cite{Henry:2023_LOSTU_space} for calibrated measurements. The system may thus be optimally solved accounting for uncertainties in both the pixel measurements ($\bSigma_{\bar{\bu}_i}$) and 3D points ($\bSigma_{\bar{\bp}_i}$), but for simplicity we proceed here by considering the pixel noise only. For isotropic noise ($\bSigma_{\bar{\bu}_i} = \sigma_u \bS^T\bS$), the pseudoinverse is analytically obtainable
\begin{equation}
    \bSigma_{\beps_i}^\dagger = \frac{\left[ \bu_i \times \right]^2 \bS^T \bS \left[ \bu_i \times \right]^2}{\sigma_u^2  \| \bu_i \|^4  \left(\bk^T \bP \bar{\bp}_i\right)^2}.
\end{equation}
One can use the factorization $\bSigma_{\beps_i}^\dagger = \bB_i^T \bB_i$, to find
\begin{equation} \label{eq:CovCholDecomp}
\bB_i = \frac{\bS \left[ \bu_i \times \right]^2}{ \sigma_u \| \bu_i \|^2 \bk^T \bP \bar{\bp}_i}.
\end{equation}
Given an initial estimate of the projection matrix $\bP$ up to scale, the following linear system can be constructed:
\begin{equation} \label{eq:DLT_weighted_Bi}
    \begin{bmatrix}
        \bB_1 \bA_1 \\
        \bB_2 \bA_2 \\
        \vdots \\
        \bB_n \bA_n
    \end{bmatrix} \text{vec}(\bP) = \mathbf{0}_{2n\times1}.
\end{equation}

The inverse covariance of the system in Eq.~\eqref{eq:DLT_weighted_Bi} is the one commonly found for the weighted least-squares,
\begin{equation} \label{eq:invcovsoln}
    \bSigma_{\text{vec}(\bP)}^{-1} = \sum_i \bA_i^T \bSigma_{\beps_i}^{\dagger} \bA_i.
\end{equation}
Furthermore, since the system in Eq.~\ref{eq:DLT_weighted_Bi} is solved through the SVD, one can rewrite Eq.~\ref{eq:invcovsoln} as
\begin{equation}
    \bSigma_{\text{vec}(\bP)}^{-1} = \bV \bD^2 \bV^T.
\end{equation}

We may solve this even more efficiently by leveraging the problem structure. To see how this works, substituting for $\bA_i$ and $\bB_i$ from Eq.~\eqref{eq:geometric_constraint} and Eq.~\eqref{eq:CovCholDecomp}, we find that
\begin{equation}
    \bB_i \bA_i  = \frac{\bS \left[ \bu_i \times \right]^2}{ \sigma_u \| \bu_i \|^2 \bk^T \bP \bar{\bp}_i} \left( \bar{\bp}^T_i \otimes [\bar{\bu}_i \times] \right) = - q_i \bS \bA_i
\end{equation}
where the scalar $q_i$ is given by
\begin{equation} \label{eq:weights}
    q_i = \frac{1}{\sigma_u \bk^T \bP \bar{\bp_i}}.
\end{equation}

The statistically optimal solution (under isotropic measurement noise) is obtained by solving the system:
\begin{equation} \label{eq:DLT_weighted_qi}
    \begin{bmatrix}
        q_1 \bS \bA_1 \\
        q_2 \bS \bA_2 \\
        \vdots \\
        q_n \bS \bA_n
    \end{bmatrix} \text{vec}(\bP) =  \mathbf{0}_{2n\times1}.
\end{equation} 
Solving the system in Eq.~\eqref{eq:DLT_weighted_qi} finds the optimal projection matrix $\bP$ and does not require knowledge of the camera calibration. In some cases, this information is sufficient and can be used to reproject points onto pixel coordinates. In some other cases, $\bK$, $\bR$, $\bolr$ can be estimated with QR factorization and $\bP$. In the PnP, however, we wish to incorporate knowledge of the camera calibration to explicitly estimate rotation and position, as will be discussed in the following section.

\subsection{Maintaining Optimality in \texorpdfstring{$SE(3)$}{SE(3)}} \label{sec:SE3}
A problem inherent to the calibrated DLT is that the projection matrix is only recovered up to scale and there is no guarantee that $\bP$ gives a rotation in $SO(3)$. Converting the optimal $\bP$ solution back to $SO(3)$ by solving the orthogonal Procrustes problem will remove the optimality of the solution. Instead, we choose to apply weights to give the \emph{weighted} orthogonal Procrustes problem, which should maintain the optimality of the solution. 

To compute the weights, first recall that the data has already been modified to respect isotropic scaling. As a first step, we need to bring back $\text{vec}(\tilde{\bP})$ and $\bSigma_{\text{vec}(\tilde{\bP})}^{-1}$ to the regular space. By applying Eq.~\ref{eq:un-normalize}, de-clamping $\bK$ from $\bP$ as defined in Eq.~\eqref{eq:projmat}, we define the matrix 
\begin{equation}
    \bM = \left( \bT_p \otimes (\bK^{-1} \bT_u^{-1}) \right).
\end{equation}
It directly follows that
\begin{align}
    \text{vec}(\acute{\bR} [\bI, -\acute{\bolr}]) &= \bM \text{vec}(\tilde{\bP}), \\
    \bSigma_{\text{vec}(\bR [\bI, -\bolr])}^{-1} &= \bM^{-T} \bSigma_{\text{vec}(\tilde{\bP})}^{-1} \bM^{-1},
\end{align}
where the acute accent $\acute{}$ refers to the quantities have not yet been scaled/orthogonalized.

Since the inverse covariance is indicative of the information in each of the pose quantities, we compute the weight matrix $\bW \in \mathbb{R}^{3\times3}$ composed of the nine first diagonal elements of $\bSigma_{\text{vec}(\bR [\bI, -\bolr])}^{-1}$. Then, solve the Procrustes problem for 
\begin{equation} \label{eq:weighted_procrustes}
    \arg \min_{\bR \in \text{SO(3)}} \| \left(\bR - \acute{\bR} \right) \odot \bW \|_F,
\end{equation}
where $\odot$ is the element-wise multiplication. We solve Eq.~\ref{eq:weighted_procrustes} by estimating a small rotation vector deviation $\delta \bphi$ from the nearest $SO(3)$ matrix to $\acute{\bR}$, such that it minimizes this cost function. This may be done in a linear system by recognizing the small angle perturbation as $(\bI - [\delta \bphi \times])$. Alternatively, this problem may be solved by using Rodrigues parameters in an iterative nonlinear least-squares ensuring $SO(3)$, where one iteration is typically enough.

$\bolr$ can be found back by applying the scale that ensures $det(\bR) = 1$. However, its elements are not transformed in a way that maintains optimality. If the best position is needed, since the rotation has been recovered optimally, one can use LOST \cite{Henry:2022_LOST} as a non-iterative, DLT-based, and optimal $n$-view triangulation algorithm which minimizes reprojection error while maintaining $\mathcal{O}(n)$ complexity. Realizing that the weights in Eq.~\ref{eq:weights} are proportional to those in Ref.~\cite{Henry:2022_LOST}, we slice the system matrix $\bA = [\bB, \bC]$ to reformulate Eq.~\ref{eq:DLT_weighted_qi} as a triangulation problem,
\begin{equation} \label{eq:LOST_weighted_qi}
    \begin{bmatrix}
        q_1 \bS \bC_1 \\
        q_2 \bS \bC_2  \\
        \vdots \\
        q_n \bS \bC_n 
    \end{bmatrix} \bt
    =  \begin{bmatrix}
        q_1 \bS \bB_1 \\
        q_2 \bS \bB_2  \\
        \vdots \\
        q_n \bS \bB_n 
\end{bmatrix} \text{vec}(\bR).
\end{equation} 
Thus, given that we already have the optimal rotation $\bR$, we can easily estimate the optimal translation $\bt$ and without recomputing the whole DLT system.

\subsection{Comparing Our Approach to LM/GN}
The Optimal Normalized DLT requires an initial guess to compute the system weights. Therefore, why use such approach over an iterative refinement like Levenberg-Marquardt (LM) or Gauss-Newton (GN)?

LM/GN computes a step to locally deviate from the initial guess. This acts as a refinement procedure and may take multiple iterations. Depending on the initial guess, the LM/GN may fall into various local minima.

In contrast, our two-shot approach only uses the initial guess to compute relative \emph{weights}: the linear system that is solved gives a completely new solution which is somewhat independent from the initial guess. The initial guess need not be very precise to compute the weights.

Different strategies could be used for the initial estimate. First, one could use a subset of the regular DLT system to solve for $\bP$, or use smaller-sized PnP methods such as P3P and P4P. Interstingly, if outlier rejection is a concern, the initial estimate could also come from a RANSAC scheme \cite{Fischler:1981_RANSAC} along with the inlier indices and be seamlessly integrated with oDLT.

\subsection{Summary: The Proposed Implementation}
We propose the optimal DLT, \texttt{oDLT} in Algorithm.~\ref{algo:odlt}. The algorithm may be written without any for loops in languages supporting linear algebra arithmetic.
\begin{algorithm}
\caption{Optimal DLT for PnP, \texttt{oDLT}} \label{algo:odlt}
\begin{algorithmic}[1]
\State Normalize the data following Eqs.~\ref{eq:normalize_meas} and \ref{eq:normalize_points}
\State Build system $\mathbf{A}$ from Eq.~\ref{eq:DLT_to_solve} with a few measurements
\State Solve for an estimate of $\mathbf{P}$ using a subset of $\mathbf{A}$ (alternatively another fast PnP method, or RANSAC \cite{Fischler:1981_RANSAC})
\State Compute the weights $q_i$ given Eq.~\ref{eq:weights} and construct the full system as in Eq.~\ref{eq:DLT_weighted_qi}
\State Solve for $\mathbf{P}$ using the full system
\State De-normalize and solve the weighted Procrustes problem as in Section \ref{sec:SE3}
\State For better position, re-triangulate using LOST in Eq.~\ref{eq:LOST_weighted_qi} (\texttt{oDLT+LOST})
\end{algorithmic}
\end{algorithm}

\section{RESULTS} \label{sec:experiments}
\subsection{Numerical Experiments}
\begin{figure*}[ht!]
    \centering
    \begin{minipage}{0.12\linewidth}
        \hfill
    \end{minipage}%
    \begin{minipage}{0.8\linewidth}
        \hfill
        \begin{minipage}{0.29\linewidth}
            \centering
            \textbf{Rotation RMSE (deg)}
        \end{minipage}%
        \hfill
        \begin{minipage}{0.29\linewidth}
            \centering
            \textbf{Position RMSE (-)}
        \end{minipage}%
        \hfill
        \begin{minipage}{0.29\linewidth}
            \centering
            \textbf{Reprojection Error (Pixel)}
        \end{minipage}
    \end{minipage}

    \begin{minipage}{0.12\linewidth}
        \raggedleft
        \subcaption{\\\textbf{Increasing $n$\\Centered}\\$\sigma_u = 1$ pixel}
        \label{fig:n_centered}
    \end{minipage}%
    \begin{minipage}{0.8\linewidth}
        \hfill
        \begin{subfigure}{0.29\linewidth}
        \centering
        \includegraphics[width=1.\linewidth]{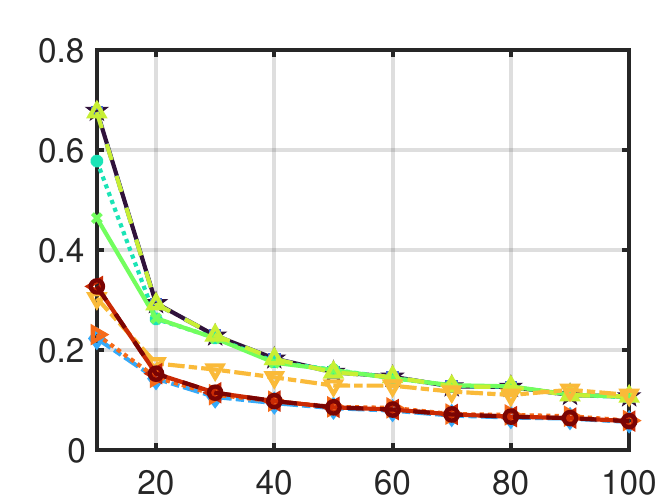}
        \end{subfigure}
        \hfill
        \begin{subfigure}{0.29\linewidth}
        \includegraphics[width=1.\linewidth]{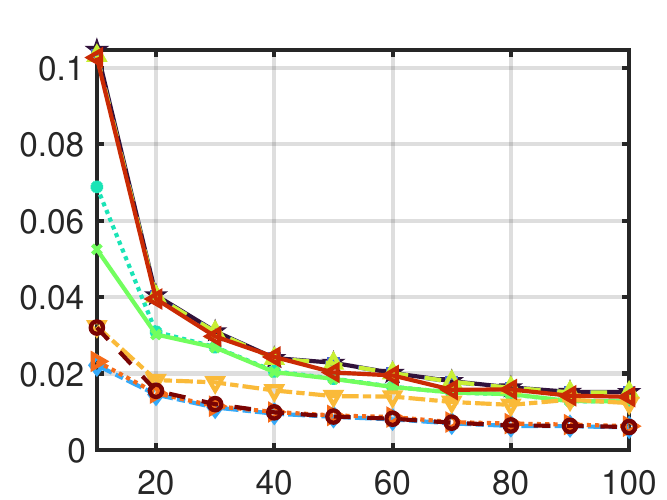}
        \end{subfigure}
        \hfill
        \begin{subfigure}{0.29\linewidth}
            \includegraphics[width=1.\linewidth]{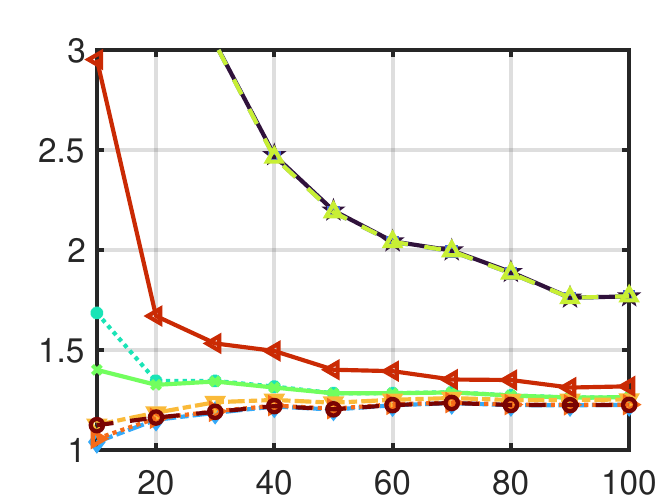}
        \end{subfigure}
    \end{minipage}

    \begin{minipage}{0.12\linewidth}
        \raggedleft
        \subcaption{\\\textbf{Increasing $n$\\Uncentered}\\$\sigma_u = 1$ pixel}
        \label{fig:n_uncentered}
    \end{minipage}%
    \begin{minipage}{0.8\linewidth}
        \hfill
        \begin{subfigure}{0.29\linewidth}
        \centering
        \includegraphics[width=1.\linewidth]{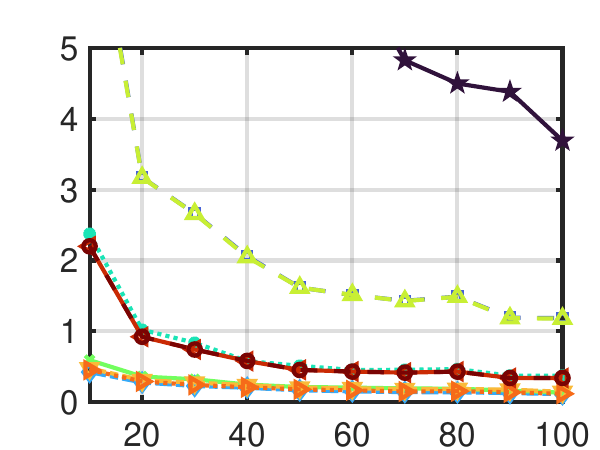}
        \end{subfigure}
        \hfill
        \begin{subfigure}{0.29\linewidth}
        \includegraphics[width=1.\linewidth]{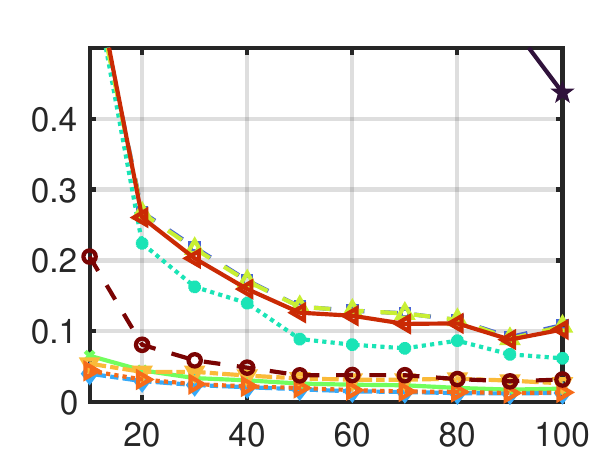}
        \end{subfigure}
        \hfill
        \begin{subfigure}{0.29\linewidth}
            \includegraphics[width=1.\linewidth]{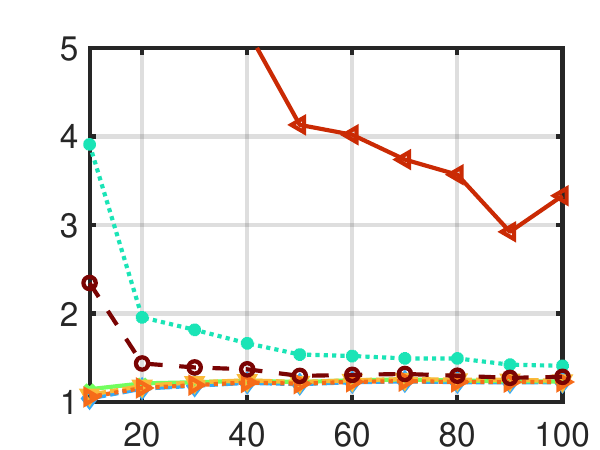}
        \end{subfigure}
    \end{minipage}

    \begin{minipage}{0.12\linewidth}
        \raggedleft
        \subcaption{\\\textbf{Increasing $\sigma_u$\\Centered}\\$n=50$}
        \label{fig:sig_centered}
    \end{minipage}%
    \begin{minipage}{0.8\linewidth}
        \hfill
        \begin{subfigure}{0.29\linewidth}
        \centering
        \includegraphics[width=1.\linewidth]{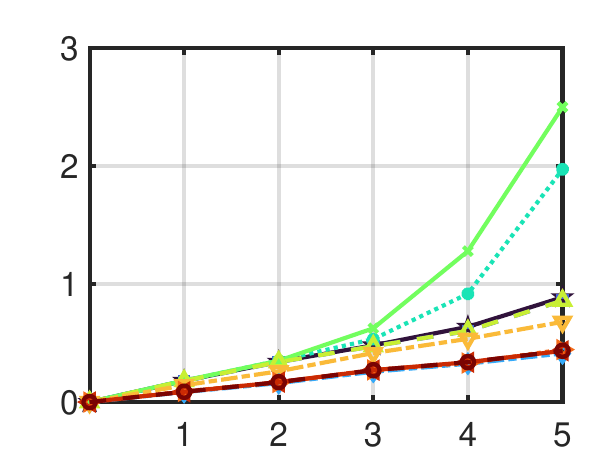}
        \end{subfigure}
        \hfill
        \begin{subfigure}{0.29\linewidth}
        \includegraphics[width=1.\linewidth]{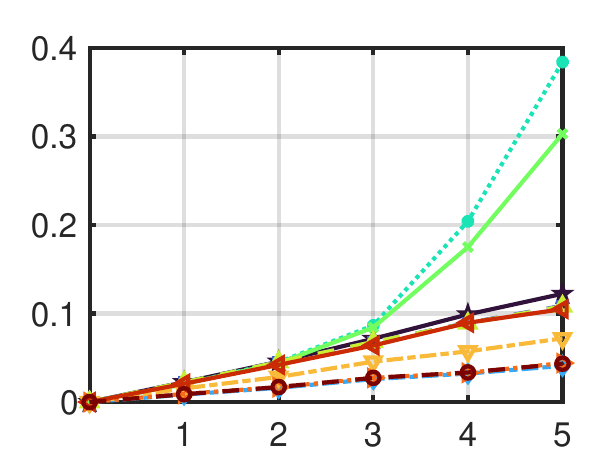}
        \end{subfigure}
        \hfill
        \begin{subfigure}{0.29\linewidth}
            \includegraphics[width=1.\linewidth]{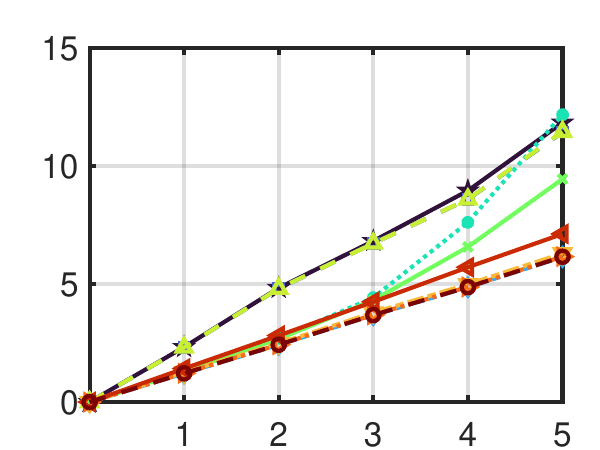}
        \end{subfigure}
    \end{minipage}

    \begin{minipage}{0.12\linewidth}
        \raggedleft
        \subcaption{\\\textbf{Increasing $\sigma_u$\\Uncentered}\\$n=50$}
        \label{fig:sig_uncentered}
    \end{minipage}%
    \begin{minipage}{0.8\linewidth}
        \hfill
        \begin{subfigure}{0.29\linewidth}
        \centering
        \includegraphics[width=1.\linewidth]{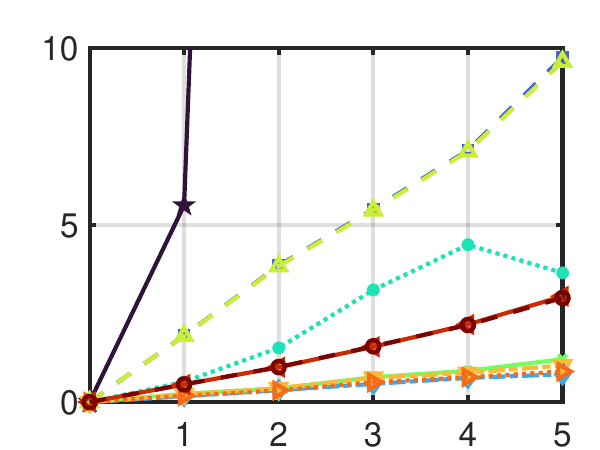}
        \end{subfigure}
        \hfill
        \begin{subfigure}{0.29\linewidth}
        \includegraphics[width=1.\linewidth]{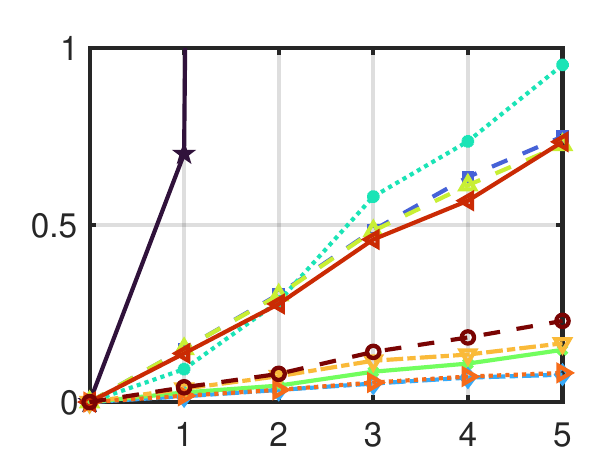}
        \end{subfigure}
        \hfill
        \begin{subfigure}{0.29\linewidth}
            \includegraphics[width=1.\linewidth]{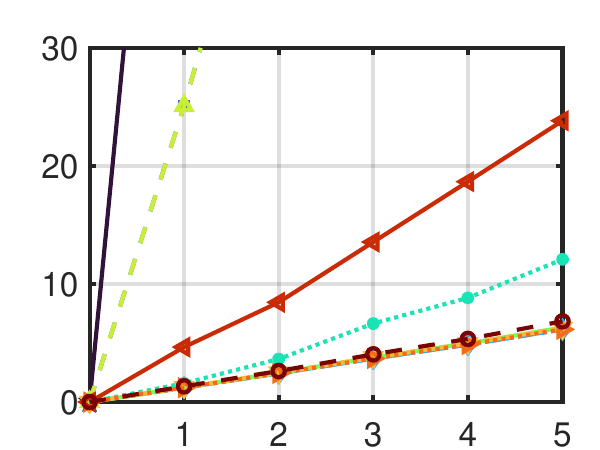}
        \end{subfigure}
    \end{minipage}
    \begin{minipage}{1.\linewidth}
        \centering
        \includegraphics[width=0.9\linewidth]{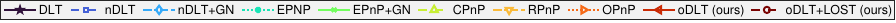}
        \label{fig:nv_legends}
    \end{minipage}
    \caption{Error metrics on different PnP methods in simulated experiments.}
    \label{fig:synthetic_results}
\end{figure*}

We compare the behavior of the DLT against different algorithms. For the DLT-based methods, we compare the regular un-normalized DLT \texttt{DLT}, the normalized DLT \texttt{nDLT}, the optimally weighted DLT \texttt{oDLT}, the optimally weighted DLT with a triangulation refinement \texttt{oDLT+LOST}, and the normalized DLT with a Gauss-Newton iteration from Ref.~\cite{Zeng:2023_CPnP} \texttt{nDLT+GN}. We compare the DLT-based methods against some of the most popular methods --- all of which are sourced from the publicly released code of the different references---, \texttt{EPnP} \cite{lepetit:2009_epnp}, \texttt{EPnP+GN} \cite{lepetit:2009_epnp}, \texttt{CPnP} \cite{Zeng:2023_CPnP}, \texttt{RPnP} \cite{li:2012_RPnP}, \texttt{OPnP} \cite{Zheng:2013_opnp}. The Gauss-Newton in \texttt{EPnP+GN} differs from a traditional Gauss-Newton as it only iterates on four parameters, and hence it is highlighted separately. 

\begin{figure}[ht!]
    \centering
    \includegraphics[width=0.7\linewidth]{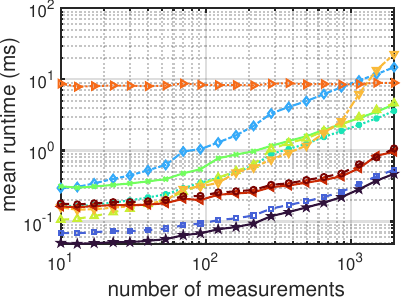}
    \caption{Mean runtimes of all the algorithms.}
    \label{fig:runtimes}
\end{figure}

\begin{table}[ht!]
    \centering
    \begin{tabular}{|c|cccc|}
    \hline
                       & RMSE $f_x$ & RMSE $f_y$ & RMSE $c_x$ & RMSE $c_y$ \\
                       \hline
         \texttt{nDLT} & 4.088 & 4.225 & 4.014 & 3.695\\
         \texttt{oDLT} & 3.917 & 4.060 & 3.823 & 3.516 \\
         \hline
    \end{tabular}
    \caption{Intrinsic parameter errors for $n$=50 (centered case).}
    \label{tab:intrinsics}
\end{table}

We only consider 2D noise and assume no covariance on the 3D points for simplicity of the framework. This assumption is consistent with all the other PnP methods listed in the prior paragraph. For full uncertainty awareness, Eq.~\ref{eq:beps_2D+3D} may be used with our method, in which case the results should be compared with MLPnP \cite{Urban:2016_MLPnP} and EPnPU \cite{Vakhitov:2021_EPNPU}.

We reproduce the same experiments as in Ref.~\cite{lepetit:2009_epnp}.  Namely, we place a camera at origin pointing upward in the $+z$ direction. The image is $640\times480$ pixels with an effective focal length of $800$ and a base isotropic noise $\sigma_u = 1$ pixel.

Starting with the centered case, we randomly sample points in the 3D box $\mathcal{B}_A \coloneqq \{(x,y,z) : x\in[-2,2], y\in[-2,2], z\in[4,8]\}$. Fig.~\ref{fig:n_centered} shows results as a function of point count. Both \texttt{DLT} and \texttt{nDLT} yield poor results, while the optimal DLT outperforms DLT significantly in rotation error and slightly in position error. \texttt{CPnP} and  \texttt{nDLT} demonstrate similar statistics. As $n$ grows, \texttt{oDLT} achieves low rotation errors, matching the optimal solutions \texttt{nDLT+GN} and \texttt{OPnP}, while \texttt{EPnP}, \texttt{EPnP+GN}, and \texttt{RPnP} perform worse than other optimal methods. Re-triangulation with LOST in \texttt{oDLT+LOST} reduces position, and thus reprojection, errors to levels of other optimal solutions. For $n\geq20$, \texttt{oDLT+LOST}, \texttt{nDLT+GN}, and \texttt{OPnP} performed best, with \texttt{oDLT+LOST} consistently the fastest.

In the uncentered case, we randomly place points in the 3D box $\mathcal{B}_B \coloneqq \{(x,y,z) : x\in[1,2], y\in[1,2], z\in[4,8]\}$. The results can be found in Fig.~\ref{fig:n_uncentered}. The \texttt{oDLT} gives a noticeably better result than the \texttt{nDLT}. \texttt{oDLT} has results that compare with \texttt{EPnP} in rotation, while \texttt{oDLT+LOST} has better position estimation. \texttt{OPnP} was the best method for estimation errors, although still the slowest by far. Methods like \texttt{OPnP}, \texttt{nDLT+GN} and \texttt{EPnP+GN} perform best, although \texttt{RPnP} exhibits a very good performance for a non-iterative method. One can see a clear improvement of the \texttt{oDLT} and \texttt{oDLT+LOST} with performance that approaches optimal PnP methods, but at a lower computational cost.

The runtimes of our implementation for increasing $n$ have been computed using MATLAB's \texttt{timeit} function, and are shown in Fig.~\ref{fig:runtimes}. We observe that our method is faster than an iterative refinement and is on par with \texttt{EPnP+GN} for small $n$. For $n\geq50$, our proposed method exhibits a faster runtime than all other proposed optimal methods. \texttt{OPnP} has a nearly flat slope on the runtime but a high constant, similar to observations in Ref.~\cite{Zheng:2013_opnp}.

If the cameras are uncalibrated, we found that \texttt{oDLT} had smaller root-mean-square error (RMSE) when estimating the calibration matrix than \texttt{nDLT}, as shown in Table~\ref{tab:intrinsics}.

\subsection{Real Data: Structure from Motion}
\begin{table*}[ht!]
\centering
\caption{Camera pose estimation performance metrics on the ETH3D structures. \texttt{nDLT+GN} is used with one iteration only. The best score is written in \textbf{bold}, second best in \underline{underline}, and third best in \textit{italic}.}
\resizebox{\linewidth}{!}{
\begin{tabular}{|l|l|ccccccccc|}
 \hline 
 Scene & Metric & nDLT & nDLT+GN & EPNP & EPnP+GN & CPnP & RPnP & OPnP & oDLT & oDLT+LOST\\ 
 \hline 
courtyard & Rot. RMSE (deg) &  0.01276 & \textbf{ 0.00480} &  0.00894 &  0.00901 &  0.01271 &  0.01332 &  \textit{0.00867} & \underline{ 0.00856} & \underline{ 0.00856} \\ 
average $n$ 3630 & Pos. RMSE (m) &  0.00416 & \textbf{ 0.00118} &  0.00257 &  0.00256 &  0.00414 &  0.00374 & \textit{ 0.00219} &  0.00370 & \underline{ 0.00182} \\ 
 & Mean Reproj. Err. (pixel) &  1.14555 & \textbf{ 0.87840} &  0.90392 &  0.90351 &  1.14389 &  0.91623 & \textit{ 0.88692} &  1.04033 & \underline{ 0.88508} \\    & Mean Run Time (ms) & \textbf{ 1.62758} & 11.87806 &  7.41003 &  9.44868 &  8.74226 & 150.15910 & 10.16712 & \underline{ 1.97541} & \textit{ 2.27364}\\ 
\hline
delivery area & Rot. RMSE (deg) &  0.01054 & \textbf{ 0.00354} &  0.00802 &  \textit{0.00796} &  0.01054 &  0.01005 &  0.00867 & \underline{ 0.00708} & \underline{ 0.00708} \\ 
average $n$ 2591 & Pos. RMSE (m) &  0.00170 & \textbf{ 0.00040} &  0.00098 &  0.00098 &  0.00169 &  0.00128 & \textit{ 0.00084} &  0.00095 & \underline{ 0.00063} \\ 
 & Mean Reproj. Err. (pixel) &  1.11030 & \textbf{ 0.87883} &  0.92440 &  0.91407 &  1.10981 &  0.96754 & \textit{ 0.89995} &  0.94413 & \underline{ 0.88454} \\    & Mean Run Time (ms) & \textbf{ 1.09411} &  8.02652 &  5.14786 &  6.50346 &  6.20529 & 69.09487 &  9.91771 & \underline{ 1.40891} & \textit{ 1.44148} \\ 
\hline
electro & Rot. RMSE (deg) &  0.01418 & \textbf{ 0.00644} &  0.00993 &  0.00964 &  0.01419 &  0.01444 & \underline{ 0.00935} & \textit{ 0.00943} &  \textit{0.00943} \\ 
average $n$ 1689 & Pos. RMSE (m) &  0.00334 & \textbf{ 0.00072} &  0.00220 &  0.00214 &  0.00334 &  0.00394 & \textit{ 0.00164} &  0.00178 & \underline{ 0.00091} \\ 
 & Mean Reproj. Err. (pixel) &  1.52473 & \textbf{ 1.06543} &  1.16191 &  1.15513 &  1.52474 &  1.24362 & \textit{ 1.11993} &  1.19064 & \underline{ 1.07927} \\    & Mean Run Time (ms) & \textbf{ 0.73605} &  4.88350 &  3.38997 &  3.90503 &  4.28200 & 22.86642 &  9.48939 & \underline{ 1.04236} & \textit{ 1.17802} \\ 
\hline
facade & Rot. RMSE (deg) &  0.01460 & \textbf{ 0.00764} &  0.01416 &  0.01429 &  0.01460 &  0.01743 &  \textit{0.01409} & \underline{ 0.00845} & \underline{ 0.00845} \\ 
average $n$ 5441 & Pos. RMSE (m) &  0.00626 & \textbf{ 0.00229} &  0.00416 &  0.00413 &  0.00626 &  0.00551 &  0.00428 & \textit{ 0.00246} & \underline{ 0.00230} \\ 
 & Mean Reproj. Err. (pixel) &  1.41559 & \textbf{ 1.00281} &  1.09949 &  1.08928 &  1.41535 &  1.10619 & \textit{ 1.05961} &  1.06064 & \underline{ 1.01049} \\    & Mean Run Time (ms) & \textbf{ 3.04381} & 18.96400 & 11.93430 & 13.60051 & 14.07206 & 311.86771 & 10.60341 & \underline{ 3.29194} & \textit{ 3.64926} \\ 
\hline
kicker & Rot. RMSE (deg) &  0.06621 & \textbf{ 0.00612} &  0.03165 & \underline{ 0.01665} &  0.06597 &  0.02171 & \textit{ 0.01820} &  0.03687 &  0.03687 \\ 
average $n$ 1834 & Pos. RMSE (m) &  0.00274 & \textbf{ 0.00022} &  0.00132 &  0.00101 &  0.00273 &  0.00314 &  0.00091 & \underline{ 0.00061} & \textit{ 0.00074} \\ 
 & Mean Reproj. Err. (pixel) &  1.94874 & \textbf{ 1.12567} &  1.22109 &  1.19992 &  1.94993 &  1.24740 & \underline{ 1.17268} &  1.54519 & \textit{ 1.18664} \\    & Mean Run Time (ms) & \textbf{ 0.86773} &  5.64179 &  3.74228 &  4.70350 &  4.67498 & 41.11273 &  9.44427 & \underline{ 1.08272} & \textit{ 1.16335} \\ 
\hline
lounge & Rot. RMSE (deg) &  0.03691 & \textbf{ 0.01550} &  0.03048 &  0.03012 &  0.03707 & \textit{ 0.02049} & \underline{ 0.01745} &  0.05019 &  0.05019 \\ 
average $n$ 555 & Pos. RMSE (m) &  0.00562 & \textbf{ 0.00164} &  0.00331 &  0.00323 &  0.00566 & \textit{ 0.00265} & \underline{ 0.00192} &  0.00658 &  0.00553 \\ 
 & Mean Reproj. Err. (pixel) &  1.89453 & \textbf{ 1.00937} &  1.04478 & \textit{ 1.03183} &  1.89616 &  1.10052 & \underline{ 1.02390} &  1.62122 &  1.03317 \\    & Mean Run Time (ms) & \textbf{ 0.28779} &  1.79089 &  1.36481 &  1.77267 &  1.52866 &  2.70590 &  8.90559 & \underline{ 0.50243} & \textit{ 0.61988} \\ 
\hline
meadow & Rot. RMSE (deg) &  0.07546 & \textbf{ 0.01507} &  0.03472 &  0.03996 &  0.07504 &  0.03488 &  \textit{0.03302} & \underline{ 0.02965} & \underline{ 0.02965} \\ 
average $n$ 449 & Pos. RMSE (m) &  0.02873 & \textbf{ 0.00231} &  0.01354 &  0.00696 &  0.02844 &  0.00714 & \textit{ 0.00617} &  0.00810 & \underline{ 0.00526} \\ 
 & Mean Reproj. Err. (pixel) &  3.09684 & \textbf{ 1.06394} &  1.71951 &  1.26611 &  3.08190 &  1.20284 & \underline{ 1.13832} &  1.51825 & \textit{ 1.19032} \\ 
  & Mean Run Time (ms) & \textbf{ 0.29725} &  1.63963 &  1.26527 &  1.66796 &  1.41671 &  3.78238 & 11.56011 & \underline{  0.63843} & \textit{ 0.64445} \\ 
\hline
office & Rot. RMSE (deg) &  0.19941 & \textbf{ 0.03828} &  0.07677 & \textit{ 0.05826} &  0.19455 &  0.09263 & \underline{ 0.04441} &  0.15968 &  0.15968 \\ 
average $n$ 448 & Pos. RMSE (m) &  0.01407 & \textbf{ 0.00164} &  0.00319 & \textit{ 0.00254} &  0.01391 &  0.00375 & \underline{ 0.00178} &  0.01294 &  0.00558 \\ 
 & Mean Reproj. Err. (pixel) &  6.11617 & \textbf{ 1.13420} &  1.22607 & \textit{ 1.20483} &  6.08461 &  1.29424 & \underline{ 1.13843} &  5.95549 &  1.27403 \\    & Mean Run Time (ms) & \textbf{ 0.21929} &  1.69544 &  1.07039 &  1.40035 &  1.42249 &  1.62861 &  8.56979 & \underline{ 0.53607} & \textit{ 0.56049} \\ 
\hline
playground & Rot. RMSE (deg) &  0.02091 & \textbf{ 0.00698} &  \textit{0.01401} &  0.01403 &  0.02059 &  0.01438 &  0.01422 & \underline{ 0.00742} & \underline{ 0.00742} \\ 
average $n$ 1399 & Pos. RMSE (m) &  0.00136 & \textbf{ 0.00033} &  0.00122 &  0.00118 &  0.00136 &  0.00172 &  0.00109 & \textit{ 0.00043} & \underline{ 0.00034} \\ 
 & Mean Reproj. Err. (pixel) &  1.69534 & \textbf{ 1.14285} &  1.37529 &  1.35338 &  1.69218 &  1.38978 &  1.24204 & \textit{ 1.16216} & \underline{ 1.14443} \\ 
  & Mean Run Time (ms) & \textbf{ 0.64196} &  4.00431 &  2.70883 &  3.19116 &  3.42564 & 18.99210 &  9.02157 & \underline{ 0.87499} & \textit{ 0.92457} \\ 
\hline
relief & Rot. RMSE (deg) &  0.01152 & \textbf{ 0.00534} &  0.00718 &  \textit{0.00711} &  0.01152 &  0.02074 &  0.00844 & \underline{ 0.00577} & \underline{ 0.00577} \\ 
average $n$ 4069 & Pos. RMSE (m) &  0.00152 & \textbf{ 0.00033} &  0.00056 & \textit{ 0.00054} &  0.00152 &  0.00146 &  0.00057 &  0.00075 & \underline{ 0.00035} \\ 
 & Mean Reproj. Err. (pixel) &  1.18986 & \textbf{ 0.90934} &  0.93664 &  0.93350 &  1.19097 &  0.98189 & \textit{ 0.92297} &  0.99499 & \underline{ 0.91408} \\ 
  & Mean Run Time (ms) & \textbf{ 1.53425} & 11.13688 &  7.18540 &  9.29458 &  9.27056 & 180.91978 & 11.04423 & \underline{ 2.25314} & \textit{ 2.29558} \\ 
\hline
terrace & Rot. RMSE (deg) &  0.00912 & \textbf{ 0.00433} &  0.00770 &  0.00768 &  0.00907 &  0.00953 &  \textit{0.00661} & \underline{ 0.00589} & \underline{ 0.00589} \\ 
average $n$ 1618 & Pos. RMSE (m) &  0.00189 & \textbf{ 0.00056} &  0.00115 &  0.00103 &  0.00188 &  0.00212 & \textit{ 0.00092} &  0.00140 & \underline{ 0.00076} \\ 
 & Mean Reproj. Err. (pixel) &  1.25564 & \textbf{ 1.07655} &  1.11690 &  1.10841 &  1.25424 &  1.19699 & \textit{ 1.09851} &  1.14432 & \underline{ 1.08186} \\    & Mean Run Time (ms) & \textbf{ 0.72005} &  3.87565 &  2.54764 &  3.10778 &  3.14275 & 18.68520 &  9.00119 & \underline{ 0.94881} & \textit{ 1.03502} \\ 
\hline
terrains & Rot. RMSE (deg) &  0.02107 & \textbf{ 0.00563} & \underline{ 0.01654} & \textit{ 0.01672} &  0.02108 &  0.01959 &  0.01786 &  0.01958 &  0.01958 \\ 
average $n$ 1534 & Pos. RMSE (m) &  0.00264 & \textbf{ 0.00019} &  0.00072 &  0.00068 &  0.00265 &  0.00078 & \underline{ 0.00057} &  0.00225 & \textit{ 0.00058} \\ 
 & Mean Reproj. Err. (pixel) &  2.04663 & \textbf{ 0.91815} &  1.03394 &  1.02347 &  2.04920 &  1.04514 & \textit{ 0.96447} &  1.45566 & \underline{ 0.93265} \\    & Mean Run Time (ms) & \textbf{ 0.65063} &  4.21566 &  2.71331 &  3.28993 &  3.56475 & 20.12814 &  9.15482 & \underline{ 0.88520} & \textit{ 0.92203} \\ 
\hline
\end{tabular}
}
\label{tab:scene_results}
\end{table*}

\begin{figure}[ht!]
    \centering
    \includegraphics[width=0.7\linewidth]{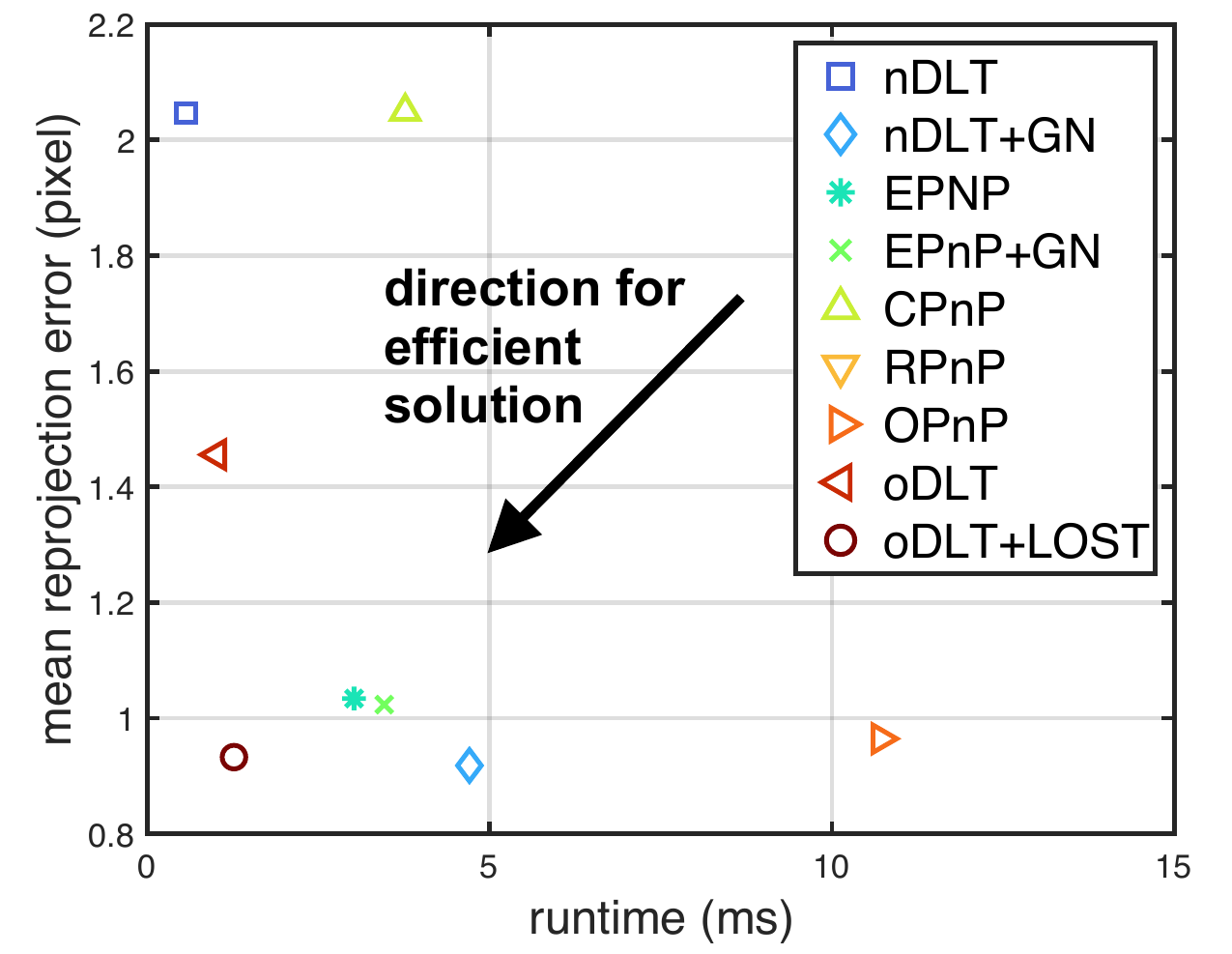}
    \caption{The scatter of time versus reprojection error on the Terrains dataset highlights a Pareto frontier. The bottom left (low runtime and low error) is the mos desirable. }
    \label{fig:pareto}
\end{figure}

Structure from motion (SfM) and visual-based reconstruction are a classic example for the use of PnP \cite{Schonberger:2016_SfMRevisited}. We select the ETH3D high-res multi-view dataset \cite{schoeps:2017_ETH3D}, which provides many scenes in different environments. ETH3D offers a COLMAP reconstruction that contains a structure of 3D points and a pose of the views. The number of points per view are high, often exceeding $n \geq 1000$. We consider the provided poses as the ground truth. We then use the provided structure to estimate the view poses and compare them against the ground truth. Some of the geometries along with images are shown in Fig.~\ref{fig:eth_visuals}, where one can observe that there are some outliers in the structure; however, no outlier rejection scheme is applied here. We provide a comparison of the errors for each method versus the ground truth along with their runtime in Table \ref{tab:scene_results}. 

First, we observe that \texttt{oDLT} is always better than \texttt{nDLT}, while having a very similar runtime. \texttt{oDLT} and \texttt{oDLT+LOST} are comparable with \texttt{EPnP} and \texttt{EPnP+GN} as they are sometimes better  (\eg delivery area, playground), and sometimes worse (\eg lounge, office). \texttt{oDLT+LOST} oftentimes shows a reduced position error lower than \texttt{EPnP+GN} and approaches what is seen with full GN methods like \texttt{nDLT+GN}. Compared to the numerical experiment, and with higher $n$ \texttt{OPnP} had a more competitive runtime, while \texttt{RPnP} usually scaled worse. Furthermore, one observes that \texttt{oDLT+LOST} gets very close to the reprojection error of iterative GN methods, but much faster. A scatter plot of time versus reprojection error is drawn in Fig.~\ref{fig:pareto} and we realize that any of the DLT-based methods are indeed Pareto-optimal. Furthermore, \texttt{oDLT+LOST} is by far the fastest method to offer low reprojection error.

\subsection{Real Data: Narrow Angle Camera}

An increasing number of space missions rely on computer vision for some critical parts of their navigation stack, with some recent examples being OSIRIS-REx~\cite{williams:2018_osiris}, DART~\cite{Rush:2023_optical}, and IM1. Many of these missions performed navigation using features of the orbserved celestial bodies. More particularly, it is quite common to use \emph{craters} for navigation around smaller bodies \cite{mcleod:2024_pnc}. Catalogs of craters are available for the Moon \cite{Robbins:2019_lunar_crater} and many other planets. Furthermore, specific algorithms exist to detect craters in an image and identify them in the aforementioned databases \cite{Woicke:2018_CraterDetection, Thrasher:2024_CraterDetection}. 

\begin{figure}[ht!]
    \centering
    \centering
    \includegraphics[width=0.99\linewidth]{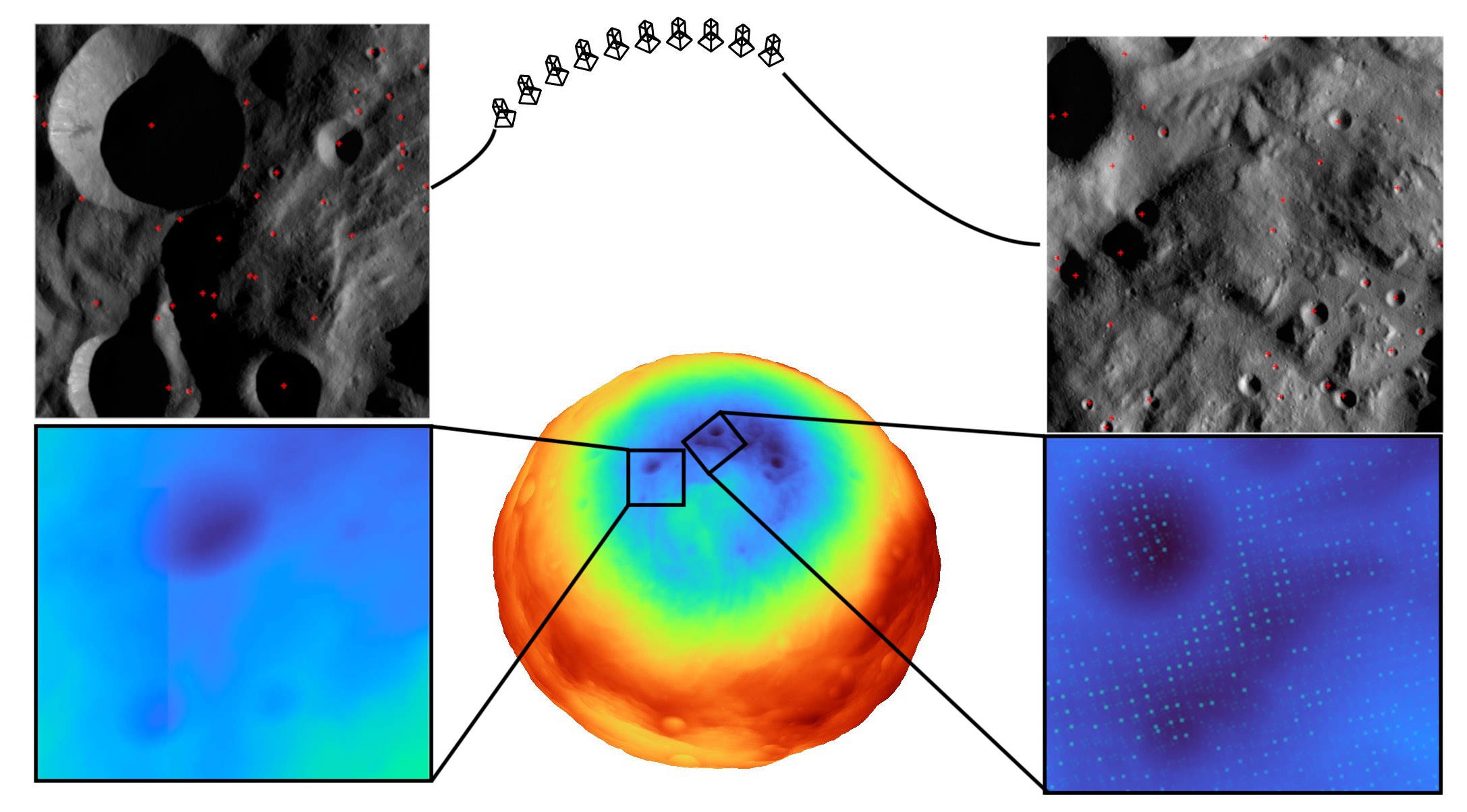}
    \caption{First (right) and last (left) images of the Dawn Vesta's \texttt{2011260\_opnav\_003} sequence with selected reconstructed cameras and identification of images on the ground-truth point cloud.}
    \label{fig:crater-nav}
\end{figure}

\begin{figure*}[ht!]
    \centering
        \begin{subfigure}{0.3\linewidth}
        \centering
        \includegraphics[width=1.\linewidth]{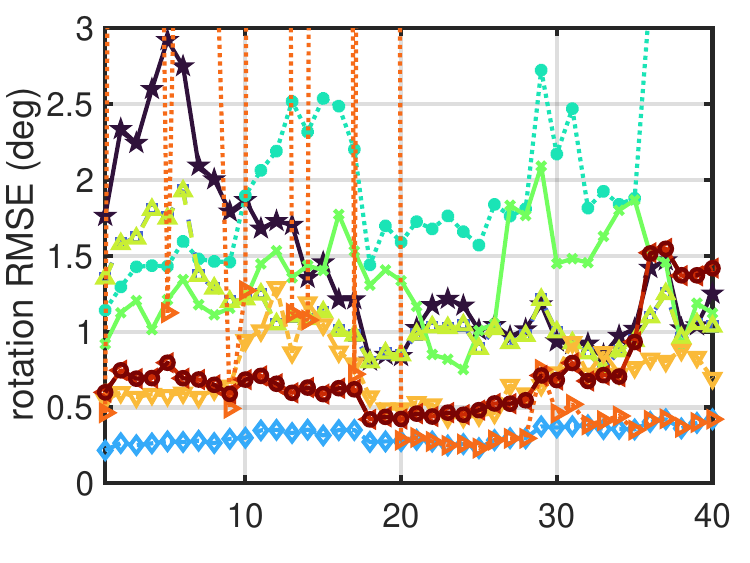}
        \end{subfigure}
        \hfill
        \begin{subfigure}{0.3\linewidth}
        \includegraphics[width=1.\linewidth]{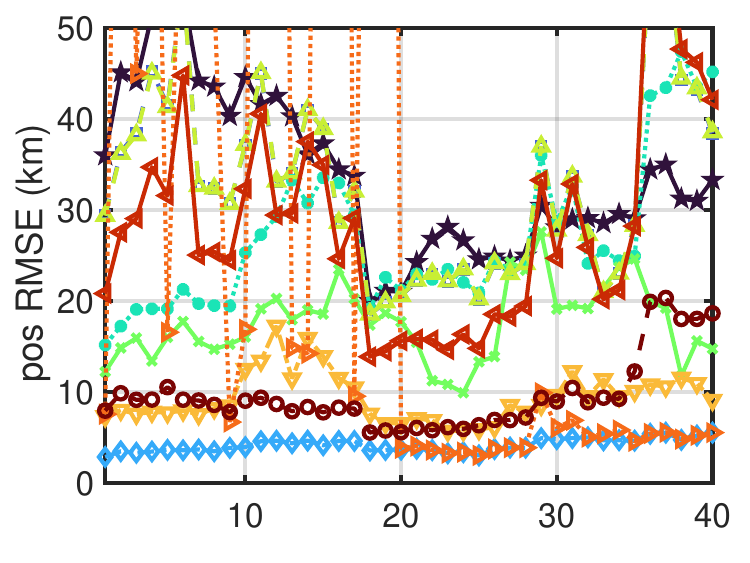}
        \end{subfigure}
        \hfill
        \begin{subfigure}{0.3\linewidth}
            \includegraphics[width=1.\linewidth]{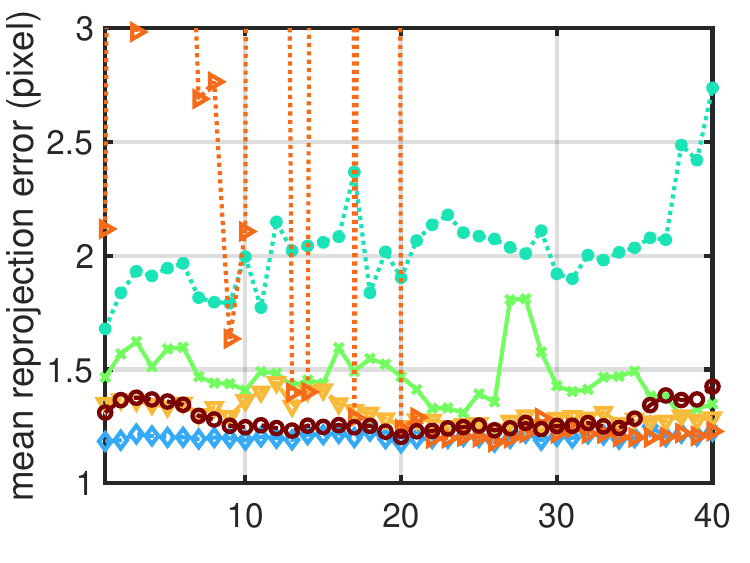}
        \end{subfigure}
    \begin{minipage}{1.\linewidth}
        \centering
        \includegraphics[width=0.9\linewidth]{legends.pdf}
    \end{minipage}
    \caption{Errors vs image number of the Dawn Vesta's \texttt{2011260\_opnav\_003} sequence.}
    \label{fig:vesta_errors}
\end{figure*}

We demonstrate an example of crater navigation using PnP with the Dawn mission \cite{russell:2011_dawn} orbiting Vesta, which has a small FOV camera of around $5^\circ$. The Astrovision dataset~\cite{Driver:2023_AstroVision} is used to obtain a clean, dense point cloud of Vesta and images from the \texttt{2011260\_opnav\_003} operational segment. This segment is composed from close-ups of the surface and highlight a rather planar geometry, as shown in Fig.~\ref{fig:crater-nav}. All images are stamped with the ground truth pose. The Vesta crater catalog proposed in Ref. \cite{liu:2018_vesta_craters} references craters with a diameter greater than 700m, and we use it to recover the longitude and latitude of craters. The 3D points of the crater centers are obtained by checking the point cloud at longitudes and latitudes. We assume that the crater identification part has already taken place and extract the corresponding measurements. We then apply a noise of 1 pixel standard deviation on the measurements. We simulate 100 Monte Carlo samples on each image of the sequence and record the errors in Fig.~\ref{fig:vesta_errors}. We observe that the data are more challenging for the PnP and that most methods have parts of the sequence where performance degrades --- images 1-20 for \texttt{OPnP}, images 10-15 for \texttt{RPnP}, and images 35-40 for \texttt{oDLT+LOST}. Despite this, \texttt{OPnP} shows significant variation on the first images and some samples did not converge at all. We also observe that the \texttt{nDLT+GN} was the most consistent with this data. In this particular geometry, multiple GN iterations in \texttt{DLT+GN} were necessary to improve the result, at the cost of runtime.

\section{CONCLUSIONS}
We derive weights to be used in the DLT which result in a statistically optimal PnP solution. Furthermore, we demonstrate that our method provides similar pose accuracy to other popular methods, but at a significantly reduced computational cost. 

\section*{ACKNOWLEDGMENT}
The authors would like to thank Lindsey Marinello and Tara Mina for their constructive feedback on this manuscript.

\bibliographystyle{IEEEtran}
\bibliography{egbib}

\end{document}